\documentclass[runningheads]{llncs}
\usepackage{amsmath,amssymb}
\usepackage{color}
\usepackage{graphicx}
\usepackage{booktabs}
\usepackage{arydshln}
\usepackage{adjustbox}
\usepackage{lipsum}
\usepackage{multirow}
\usepackage{multicol}
\usepackage[square,sort,comma,numbers]{natbib}
\usepackage{subcaption}
\usepackage{hyperref}
\usepackage[noabbrev, capitalize]{cleveref}
\usepackage{arydshln}
\usepackage{blindtext}
\usepackage{enumitem}
\usepackage{tumabbrev}
\usepackage{siunitx}
\DeclareSIUnit\px{px}
\usepackage{glossaries}
\newacronym[]{gls:RS}{RS}{remote sensing}
\newacronym[]{gls:GSD}{GSD}{ground sampling distance}
\newacronym[]{gls:CNN}{CNN}{convolutional neural network}
\newacronym[]{gls:CRF}{CRF}{conditional random field}
\newacronym[]{gls:SGD}{SGD}{Stochastic Gradient Descent}
\newacronym[]{gls:FCNN}{FCNN}{fully convolutional neural network}
\newacronym[]{gls:DCNN}{DCNN}{deep convolutional neural network}
\newacronym[]{gls:ACNN}{ACNN}{atrous convolutional neural network}
\newacronym[longplural={oriented bounding boxes}]{gls:OBB}{OBB}{oriented bounding box}
\newacronym[longplural={horizontal bounding boxes}]{gls:HBB}{HBB}{horizontal bounding box}
\newacronym[]{gls:UAV}{UAV}{unmanned aerial vehicles}
\newacronym[]{gls:miou}{mIoU}{mean intersection over union}
\newacronym[]{gls:iou}{IoU}{intersection over union}
\newacronym[]{gls:map}{mAP}{mean average precision}
\newacronym[]{gls:ar}{AR}{average recall}
\newacronym[]{gls:GFLOP}{GFLOP}{giga floating point operation}
\newacronym[]{gls:RCNN}{RCNN}{Region-based convolutional neural network}
\newacronym[]{gls:RPN}{RPN}{region proposal network}
\newacronym[]{gls:ROI}{ROI}{region of interest}
\newacronym[]{gls:FPN}{FPN}{feature pyramid network}
\newacronym[]{gls:HOG}{HOG}{histogram of oriented gradients}
\newacronym[]{gls:DAB}{DAB}{domain adapter block}
\newacronym[]{gls:NMS}{NMS}{non-maximum suppression}
\newacronym[]{gls:FPS}{FPS}{frame per second}
\newacronym[]{gls:SSD}{SSD}{single shot detector}
\newacronym[]{gls:ICN}{ICN}{image cascade network}
\newacronym[]{gls:DIN}{DIN}{deformable inception network}
\newacronym[]{gls:R-RPN}{R-RPN}{multi-scale rotational region-proposal network}
\newacronym[]{gls:R-ROI}{R-ROI}{multi-scale rotational region of interest network}
\newacronym[]{gls:R-NMS}{R-NMS}{rotational non-maximum suppression}
\newacronym[]{gls:OHEM}{OHEM}{online hard example mining}
\newglossaryentry{gls:FastRCNN}{name={Fast-\gls{gls:RCNN}}, description={}}
\newglossaryentry{gls:FasterRCNN}{name={Faster-\gls{gls:RCNN}}, description={}}
\newglossaryentry{gls:MaskRCNN}{name={Mask-\gls{gls:RCNN}}, description={}}
\glsdisablehyper
\defglsentryfmt{
	\ifglsused{\glslabel}{
		\glsgenentryfmt
	}{
	\emph{\glsgenentryfmt}}}
\newcommand{\overviewfigure}{
	\begin{figure}[t]
		\centering
		\includegraphics[width=0.99\textwidth]{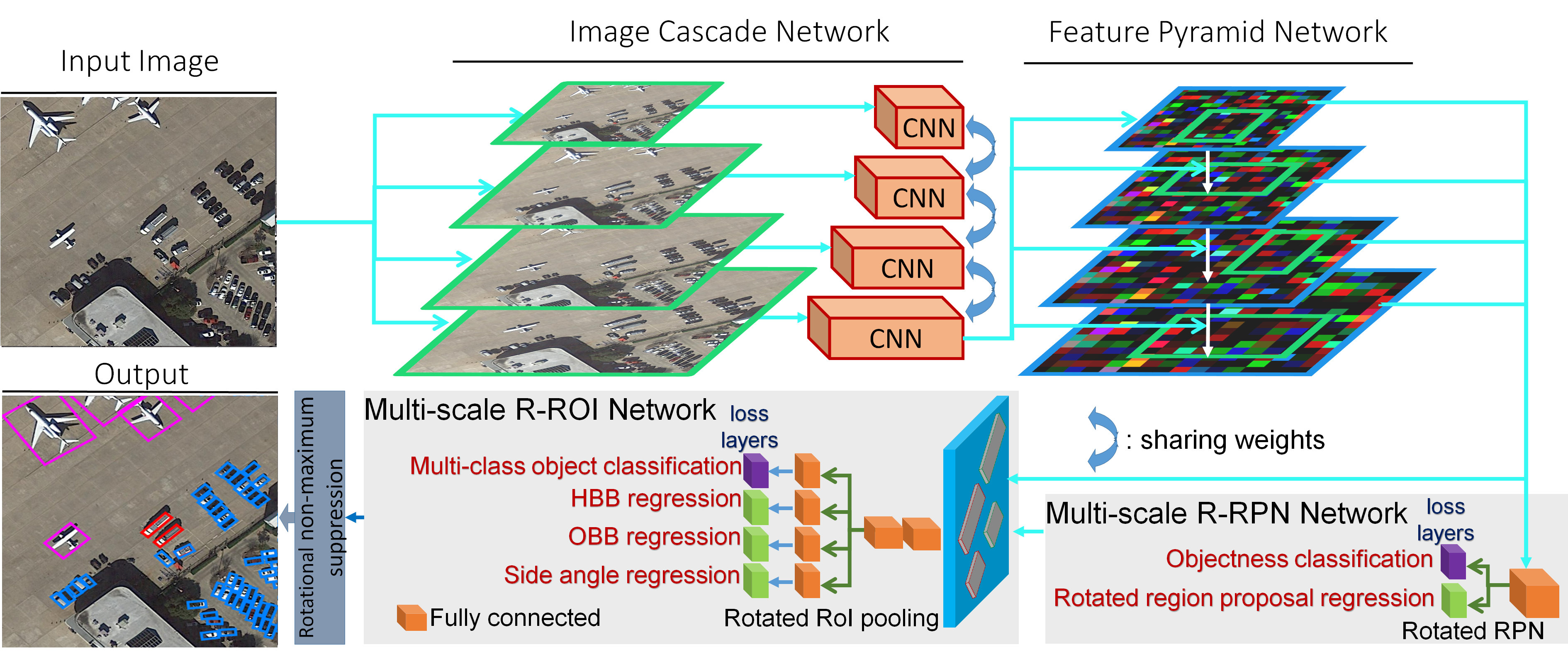}
		\caption{Overview of our algorithm for (non-)rotated multi-class object detection.}
		\label{fig:overflow}
	\end{figure}
}

\newcommand{\cascadefigure}{
	\begin{figure}[t]
		\centering
		\includegraphics[width=0.95\textwidth]{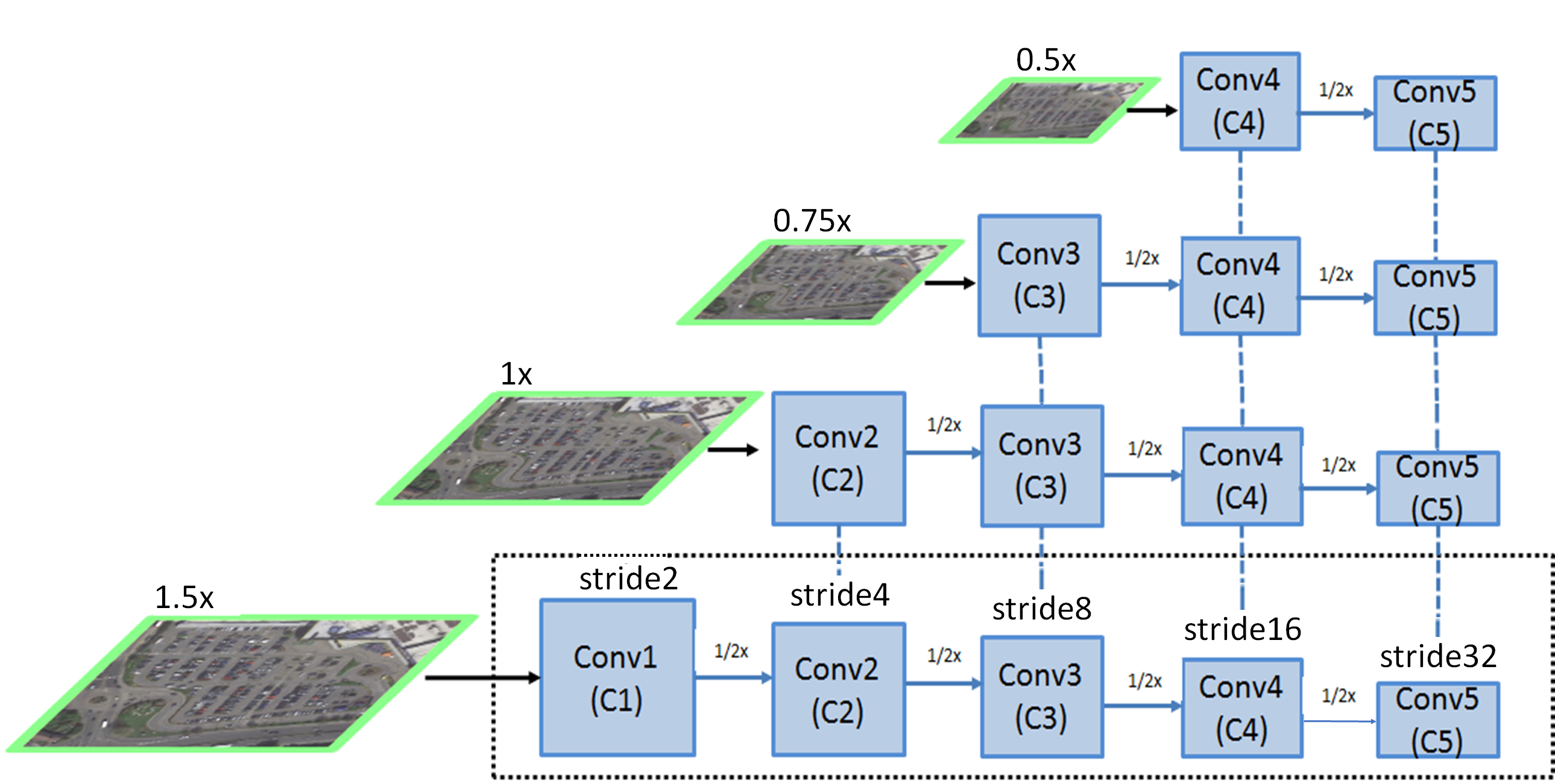}
		\caption{Illustration of the Image Cascade Network (ICN). Input images are first up- and down-sampled. Then they are fed into different CNN cascade levels.}
        \label{fig:icn}
	\end{figure}
 }
 \newcommand{\fpmidnfigure}{
  	\begin{figure}[h]
		\centering
		\includegraphics[width=\textwidth]{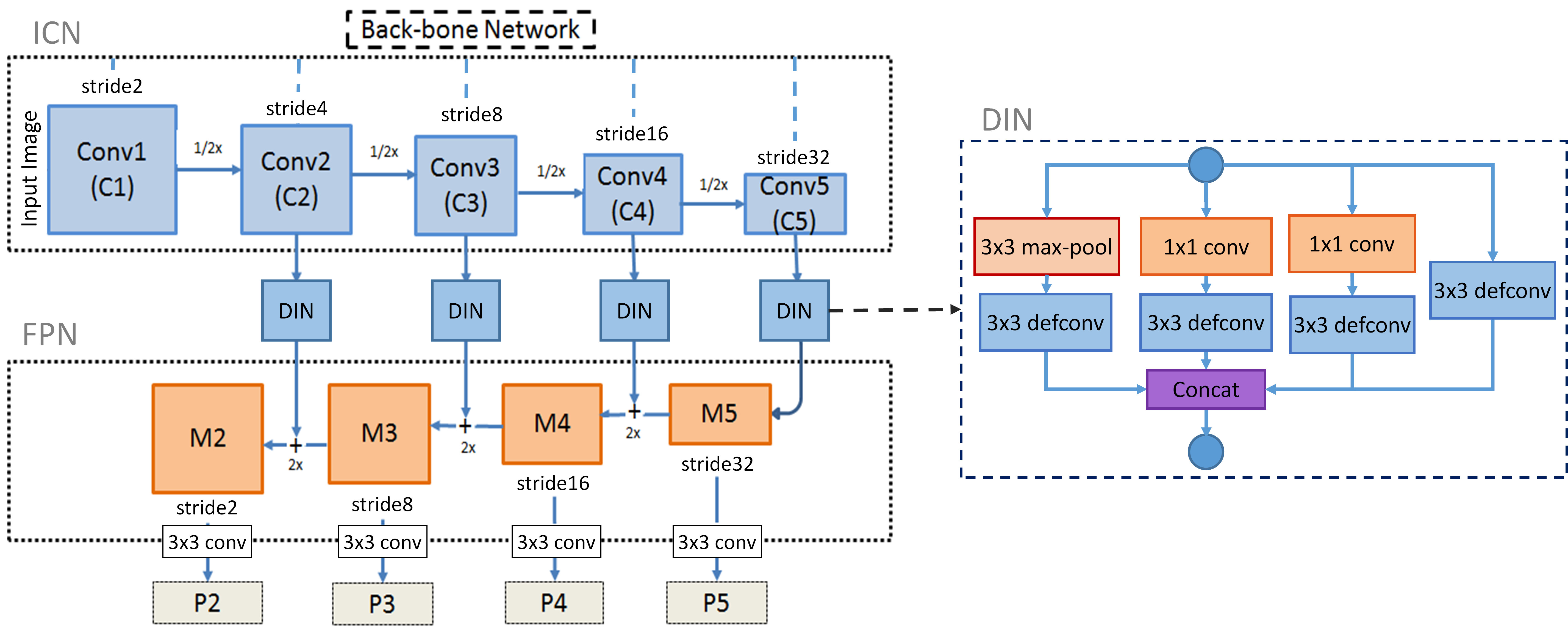}
		\caption{Illustration of the ICN and FPN subnetworks with deformable inception network (DIN). DIN is the modified Inception block to learn features of objects including geometrical features in flexible kernel sizes with stride 1. ``defconv'' stands for deformable convolution.}
        \label{fig:FPN}
	\end{figure}
}

 \newcommand{\dotaclassesfigure}{
	\begin{figure}[t]
		\centering
		\includegraphics[width=\linewidth]{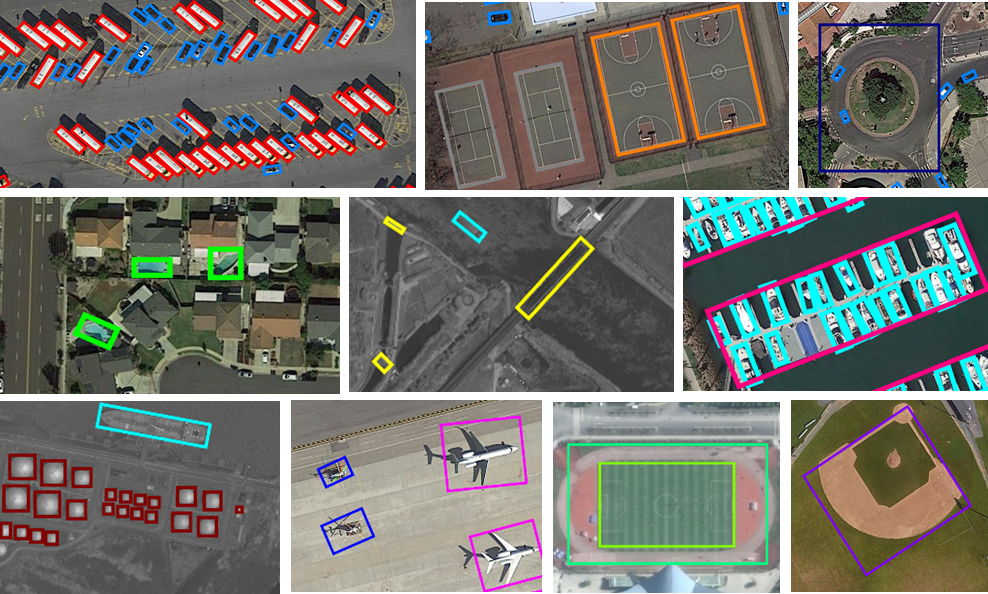}
		\caption{Sample OBB predictions in the DOTA test set.}
		\label{fig:DOTA-classes}
	\end{figure}
}

 \newcommand{\dotavehiclesfigure}{
\begin{figure}[t]
		\centering
		\includegraphics[width=0.9\linewidth]{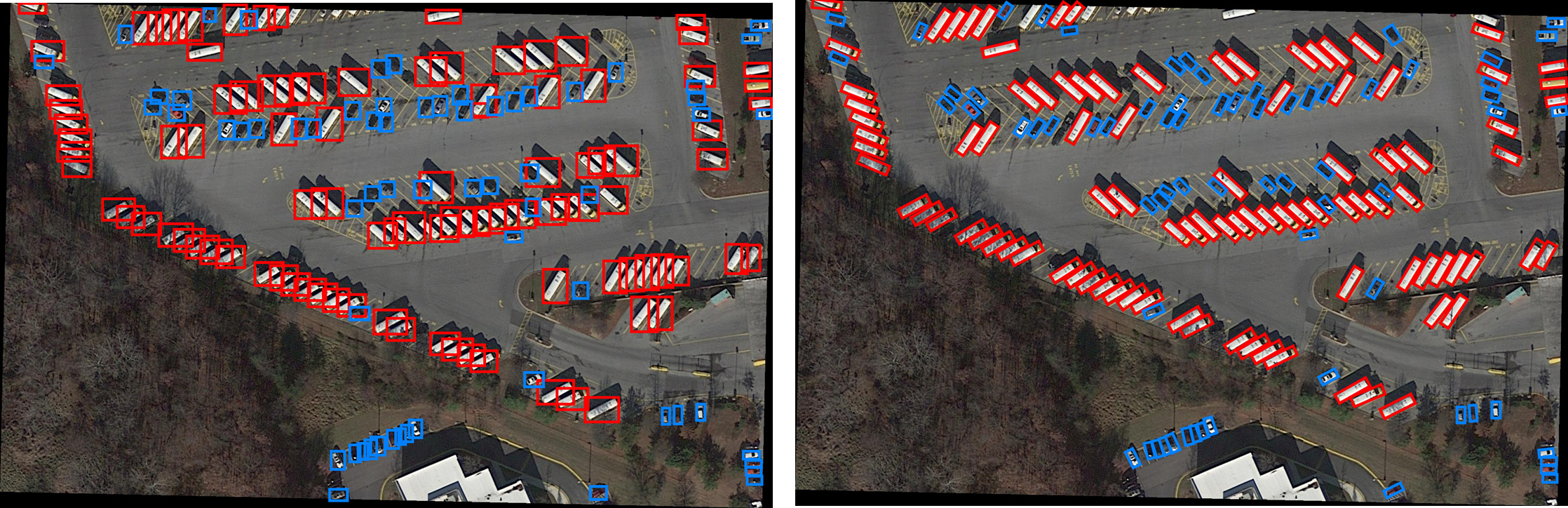}
		\caption{Outputs of HBB (left) and OBB (right) prediction on an image of DOTA.}
		\label{fig:dotavehicles}
	\end{figure}
}

\newcommand{\ucasnwpfigure}{
    	\begin{figure}[t]
		\centering
		\includegraphics[width=\linewidth]{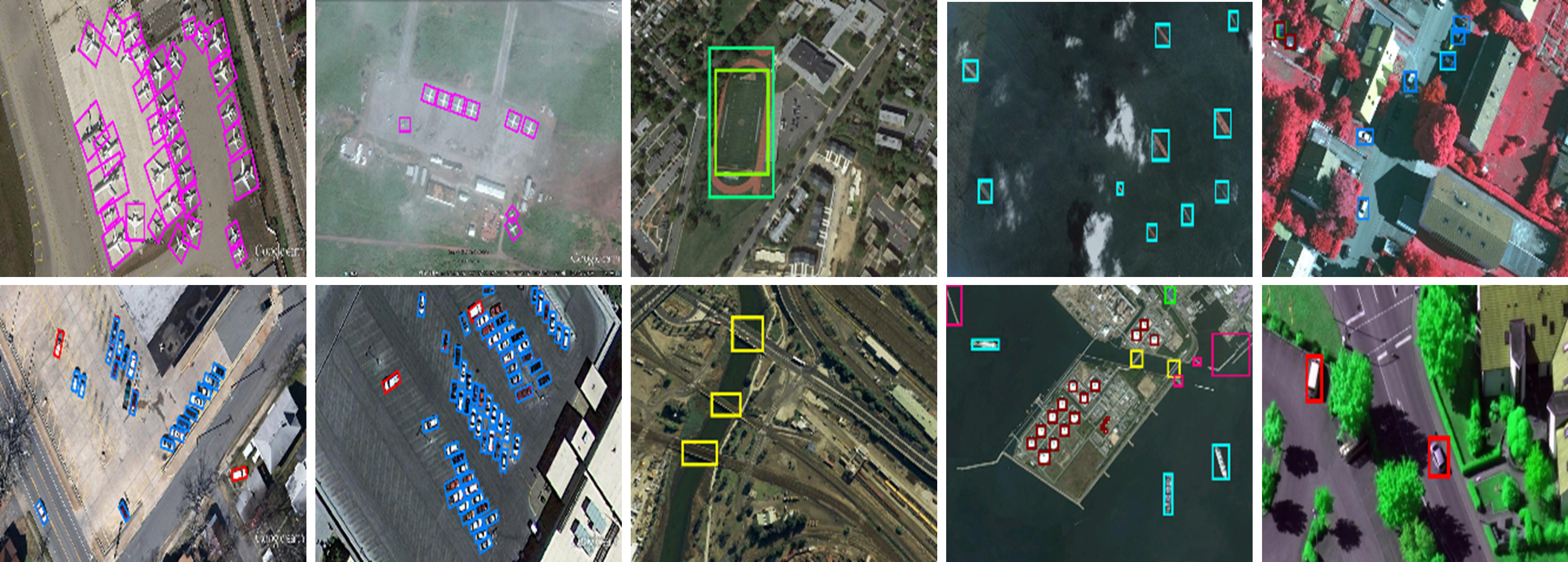}
		\caption{Sample outputs of our algorithm on the NWPU VHR-10 (three right columns -- different camera sensors) and UCAS-AOD (two left columns -- different weather conditions, camera angles, and GSDs) datasets.}
		\label{fig:nw}
	\end{figure}
}

\newcommand{\fpanalysisfigure}{
\begin{figure*}
		\centering
		\begin{subfigure}[b]{0.35\textwidth}
			\includegraphics[width=1\linewidth]{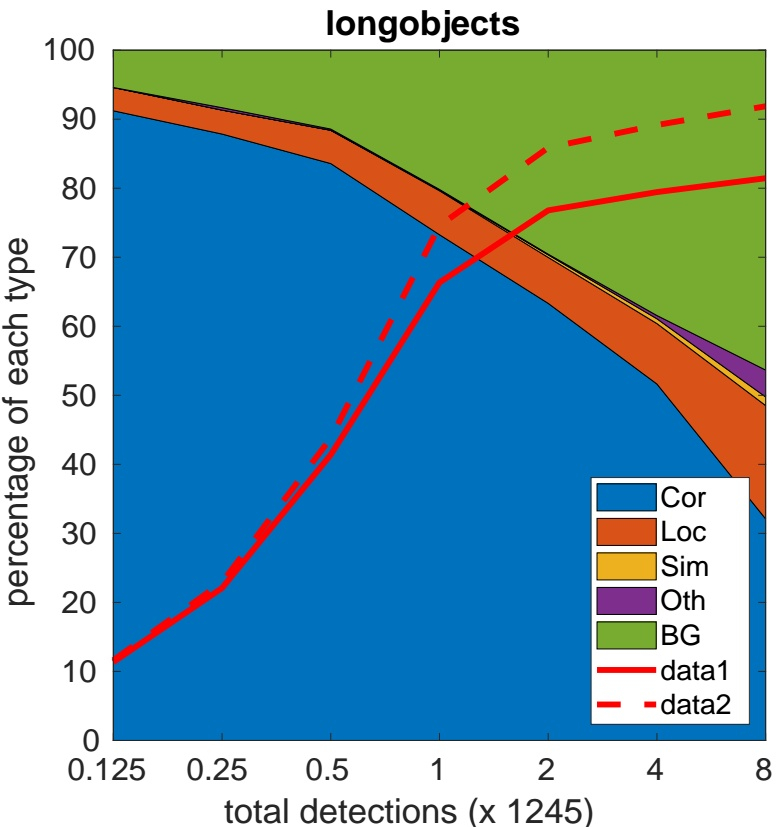}
		\end{subfigure}
		\begin{subfigure}[b]{0.35\textwidth}
			\includegraphics[width=1\linewidth]{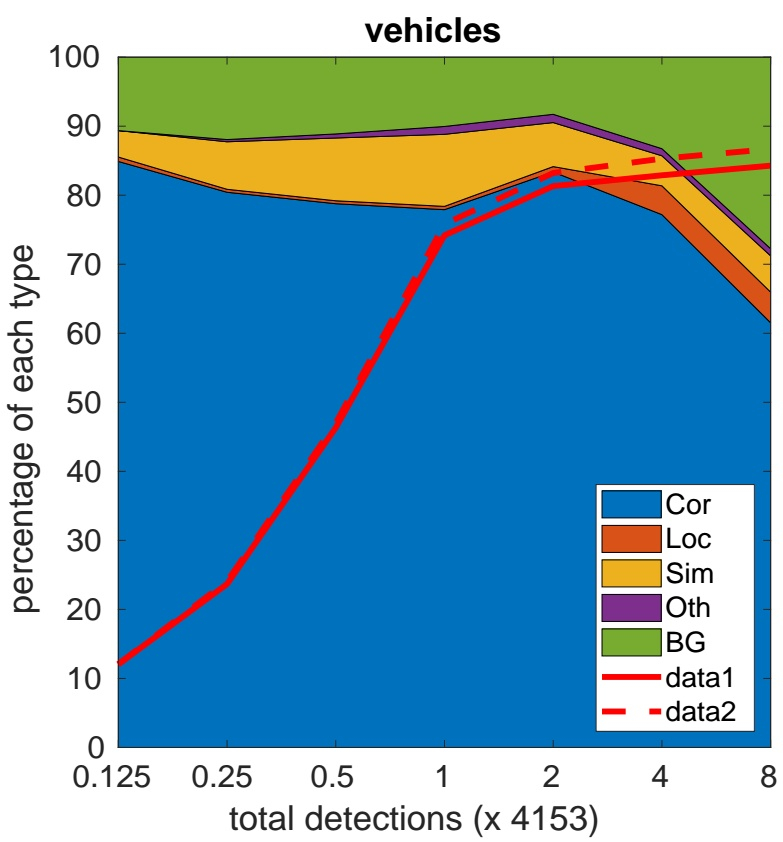}
		\end{subfigure}
		\caption{False positive trends. Stacked area plots show the fraction of each type of false positive by increasing the number of detections; line plots show recall for the weak localization with more 10\% overlap with ground truth (dashed line) and the strong one with more than 50\% overlap (solid line). Cor: correct, Loc: localization, Sim:similar classes, Oth: other reasons, BG: background.}
		\label{fig:fps}
	\end{figure*}
}
\newcommand{\icntable}{
	\begin{table*}[t]
		\centering
		\caption{Evaluation of (1) the impact of ICN with different cascade levels, (2) the effect of the backbone network (ResNet50/101, ResNeXt101), and (3) the influence of the number of proposals for the OBB prediction task. The models were trained on the DOTA training set and results are on the validation set.}
		\label{tab:cascade}
		\begin{tabular}{c|c|c|c|c}
			Cascade level & Proposals & Backbone & OHEM 
			& mAP (\%) \\
			\midrule
			1                  &300  & ResNet-50 & ---         
			& 63.35\\
			1                  &300  & ResNet-50 &  \checkmark 
			& 64.61\\
			1                  &300  & ResNet-101&  \checkmark 
			& 65.37\\
			$[1.5,1]$            & 300 & ResNet-101&  \checkmark
			& 67.32\\
			$[1.5,1,0.75]$       & 300 & ResNet-101&  \checkmark
			& 68.06\\
			$[1.5,1,0.75,0.5]$   & 300 & ResNet-101&  \checkmark
			& 68.17\\
			$[1.5,1,0.75,0.5]$   & 300 & ResNeXt-101&  \checkmark
			& 68.09\\
			$[1.5,1,0.75,0.5]$  & 2000& ResNet-101&  \checkmark
			& \textbf{68.29}\\						
			$[1.75,1.5,1,0.75]$  & 2000& ResNet-101&  \checkmark
			& 67.36\\						
			$[2,1.5,1.5,1,0.75]$ & 2000& ResNet-101&  \checkmark
			& 66.86\\							
		\end{tabular}
	\end{table*}
}

\newcommand{\dintable}{
		\begin{table*}[t]
		\centering
		\caption{Evaluation of employing DIN after certain residual blocks $C_i$ with and without deformable convolutions on the validation set of DOTA.}
		\label{tab:din}
		\begin{tabular}{c|c|c}
			DIN & Def. conv. & mAP (\%) \\
			\midrule
			-  & - &65.97\\
			C4  & - &66.24\\
			C5  & - &66.28\\						
			C4---C5  & - & 66.41\\					
			C3---C5 & - & 66.75\\							
			C2---C5 & - & 67.47\\
			C2---C5 & \checkmark & \textbf{68.17}\\		
		\end{tabular}
	\end{table*}
}

\newcommand{\rpnroitable}{
	\begin{table*}[t]
    \centering
		\caption{Evaluation of (1) the impact of rotated RPN and RoI and (2) the effect of the loss functions enforcing the rectangularity of the bounding boxes.}
		\label{tab:r}
		\begin{tabular}{c|c|c}
			Angle Loss functions & Rotated BBs in RPN \& RoI & mAP (\%) \\
			\midrule
			-  & - & 64.27\\
			- & \checkmark & 65.67\\
            \midrule
			 Tangent L1  & \checkmark & 66.91\\
			 Smooth L1  & \checkmark & 67.41\\
			 L2 & \checkmark & \textbf{68.17}\\	
		\end{tabular}
	\end{table*}
}

\newcommand{\HBBbenchmarktable}{
		\begin{table*}[t]\label{table:hbbclsresults}
		\centering
		\caption{Quantitative comparison of the baseline and our method on the HBB task in test set of DOTA dataset. FR-H stands for Faster R-CNN\cite{fasterrcnnNIPS2015} trained on HBB. TV stands for `trainval' and T for `train' subsets.}
		\begin{adjustbox}{width=1\textwidth}
			\label{tab:benchmarkhbb}
			\begin{tabular}{c|c|c||ccccccccccccccc}
				method& data  &mAP    &plane       &BD   &bridge&GTF  &SV   &LV   &ship &TC   &BC   &ST   &SBF  &RA   &harbor&SP   &HC \\
				\toprule\toprule
				Yolov2-\cite{redmon2017yolo9000} & TV &39.20&76.90&33.87&22.73&34.88&38.73&32.02&52.37&61.65&48.54&33.91&29.27&36.83&36.44&38.26&11.61\\
				R-FCN\cite{NIPS2016_6465} & TV &52.58 &	81.01 &	58.96 &	31.64 &	58.97 &	49.77 &	45.04 &	49.29 &	68.99 &	52.07& 	67.42 &	41.83 &	51.44 &	45.15 &	53.3 &	33.89\\
				SSD\cite{DBLP:conf/eccv/LiuAESRFB16} & TV &29.86 &	57.85 &	32.79 &	16.14 &	18.67 &	0.05 &	36.93 &	24.74 &	81.16 &	25.1 &	47.47 &	11.22 &	31.53 &	14.12 &	9.09 &	0.0\\
				FR-H\cite{fasterrcnnNIPS2015} & TV&60.64&80.32&77.55&32.86&68.13&53.66&52.49&50.04&90.41&75.05&59.59&57.00&49.81&61.69&56.46&41.85\\
				ours & T    &70.54&89.54&73.48&51.96&70.33&73.39&67.91&78.15&90.39&78.73&78.48&51.02&59.41&73.81&69.00&52.59\\
				ours & TV   &\textbf{72.45}&\textbf{89.97}&\textbf{77.71}&\textbf{53.38}&\textbf{73.26}&\textbf{73.46}&\textbf{65.02}&\textbf{78.22}&\textbf{90.79}&\textbf{79.05}&\textbf{84.81}&\textbf{57.20}&\textbf{62.11}&\textbf{73.45}&\textbf{70.22}&\textbf{58.08}\\	
			\end{tabular}
		\end{adjustbox}
	\end{table*}
}

\newcommand{\OBBbenchmarktable}{
	\begin{table*}[t]
		\centering
		\caption{Quantitative comparison of the baselines and our method on the OBB prediction task in test set of DOTA dataset. Abbreviations are the same as in Table \ref{tab:benchmarkhbb}.  Note that only FR-O\cite{fasterrcnnNIPS2015} is trained with OBB.}
		\begin{adjustbox}{width=1\textwidth}
			\label{tab:benchmarkobb}
			\begin{tabular}{c|c|c||ccccccccccccccc}
				method& data   &mAP    &plane       &BD   &bridge&GTF  &SV   &LV   &ship &TC   &BC   &ST   &SBF  &RA   &harbor&SP   &HC \\
				\toprule\toprule
				Yolov2-\cite{redmon2017yolo9000} & TV &
				25.49 &	52.75 &	24.24 &	10.6 &	35.5 &	14.36 &	2.41 &	7.37 &	51.79 &	43.98 &	31.35 &	22.3 &	36.68 &	14.61 &	22.55 &	11.89\\
				R-FCN\cite{NIPS2016_6465} & TV       &
				30.84 &	39.57 &	46.13 &	3.03 &	38.46 &	9.1 &	3.66 &	7.45 &	41.97 &	50.43& 	66.98 &	40.34 &	51.28 &	11.14 &	35.59 &	17.45\\
				SSD\cite{DBLP:conf/eccv/LiuAESRFB16} & TV &
				17.84  &	41.06  &	24.31  &	4.55  &	17.1  &	15.93  &	7.72  &	13.21  & 39.96  & 12.05  &	46.88  & 9.09  &	30.82  & 1.36  &	3.5  &	0.0\\
				FR-H\cite{fasterrcnnNIPS2015} & TV  &39.95 &	49.74 &	64.22 &	9.38 &	56.66 &	19.18 &	14.17 &	9.51 &	61.61 &	65.47 &	57.52 &	51.36 &	49.41 &	20.8 &	45.84 &	24.38\\
				FR-O\cite{fasterrcnnNIPS2015} &TV&54.13  &		79.42  &	\textbf{	77.13  }&		17.7  &		64.05  &		35.3  &		38.02  &		37.16  &		89.41  &		69.64  &	59.28  &		50.3  &		52.91  &		47.89  &		47.4  &	46.3\\
				R-DFPN\cite{D-FPN} & TV &57.94  &		80.92  &	65.82&		33.77  &		58.94  &		55.77  &		50.94  &		54.78  &		90.33  &		66.34  &	68.66  &		48.73  &		51.76  &		55.10  &		51.32  &	35.88\\
				Yang et al.\cite{XueYang2018} & TV &62.29  &		81.25 &	71.41&		36.53  &		67.44  &		61.16  &	50.91	  &		56.60  &		90.67  &		68.09  &	72.39  &	55.06  &	55.60   &		62.44  &	53.35  &\textbf{51.47}\\
				ours & T &64.98
				& 81.24& 68.74
				& 43.36& 61.07
				& \textbf{65.25}&67.72
				& 69.20& 90.66
				& 71.47& 70.21
				& \textbf{55.41}& 57.28
				& 66.49& 61.3&45.27\\
				ours & TV &\textbf{68.16}
				& \textbf{81.36}& 74.30
				& \textbf{47.70}& \textbf{70.32}
				& 64.89& \textbf{67.82}
				& \textbf{69.98}& \textbf{90.76}
				& \textbf{79.06}& \textbf{78.20}
				& 53.64& \textbf{62.90}
				& \textbf{67.02}& \textbf{64.17}&50.23\\
			\end{tabular}
		\end{adjustbox}
	\end{table*}
}

\newcommand{\ucasnwptable}{
  \begin{table*}[b!]
    \centering
    \caption{Comparison of results on NWUH VHR-10 and UCAS-AOD
      datasets.}
    \label{tab:generalization}
    \begin{tabular}{c|c|c||c}
      method& train data & test data &mAP\\
      \midrule\midrule
      Cheng et al.\cite{Chengnwhu} & NWUH VHR-10 & NWUH VHR-10 & 72.63\\
      ours & NWUH VHR-10 & NWUH VHR-10 & 95.01\\ \cdashline{1-1}
      ours & DOTA & NWUH VHR-10 & 82.23\\
      \midrule
      Xia et al.\cite{dota}  & UCAS-AOD & UCAS-AOD & 89.41\\
      ours & UCAS-AOD & UCAS-AOD & 95.67\\ \cdashline{1-1}
      ours & DOTA & UCAS-AOD & 86.13\\
    \end{tabular}
  \end{table*}
}

\begin{document}\sloppy

\title{Towards Multi-class Object Detection in Unconstrained Remote Sensing Imagery}
\titlerunning{Towards Multi-class Object Detection in Unconstrained Aerial Imagery}

\author{Seyed~Majid~Azimi\inst{1,2^{*}}\orcidID{0000-0002-6084-2272} \and
Eleonora~Vig\inst{1}\orcidID{0000-0002-7015-6874} \and
Reza~Bahmanyar\inst{1}\orcidID{0000-0002-6999-714X} \and
Marco~K\"orner\inst{2}\orcidID{0000-0002-9186-4175} \and
Peter~Reinartz\inst{1}\orcidID{0000-0002-8122-1475}}

% for the editors' attention
%\index{Azimi, Seyed Majid}

\authorrunning{S. Azimi et al.}

\institute{
German Aerospace Center, Remote Sensing Technology Institute, Germany\and
Technical University of Munich, Chair of Remote Sensing, Munich, Germany\\
{*}Corresponding author: \email{seyedmajid.azimi@dlr.de}
}
\maketitle
\begin{abstract}
Automatic multi-class object detection in remote sensing images in unconstrained scenarios is of high interest for several applications including traffic monitoring and disaster management.
		The huge variation in object scale, orientation, category, and complex backgrounds, as well as the different camera sensors pose great challenges for current algorithms.
        In this work, we propose a new method consisting of a novel joint image cascade and feature pyramid network with multi-size convolution kernels to extract multi-scale strong and weak semantic features.
		These features are fed into rotation-based region proposal and region of interest networks to produce object detections.
        Finally, rotational non-maximum suppression is applied to remove redundant detections.
        During training, we minimize joint horizontal and oriented bounding box loss functions, as well as a novel loss that enforces oriented boxes to be rectangular.
		Our method achieves 68.16\% mAP on horizontal and 72.45\% mAP on oriented bounding box detection tasks on the challenging DOTA dataset, outperforming all published methods by a large margin ($+6$\% and $+12$\% absolute improvement, respectively).
        Furthermore, it generalizes to two other datasets, NWPU VHR-10 and UCAS-AOD, and achieves competitive results with the baselines even when trained on DOTA.
		Our method can be deployed in multi-class object detection applications, regardless of the image and object scales and orientations, making it a great choice for unconstrained aerial and satellite imagery.
\end{abstract}
\section{Introduction}
The recent advances in~\gls{gls:RS} technologies have eased the acquisition of very high-resolution multi-spectral satellite and aerial images.
Automatic~\gls{gls:RS} data analysis can provide an insightful understanding over large areas in a short time.
In this analysis, multi-class object detection (\eg vehicles, ships, airplanes, etc.) plays a major role.
It is a key component of many applications such as traffic monitoring, parking lot utilization, disaster management, urban management, search and rescue missions, maritime traffic monitoring and so on. 
    Object detection in~\gls{gls:RS} images is a big challenge as the images can be acquired with different modalities (\eg panchromatic, multi- and hyper-spectral, and Radar) with a wide range of ~\gls{gls:GSD} \eg from \SI{10}{\cm} to \SI{30}{\m}.
    Furthermore, the objects can largely vary in scale, size, and orientation.
    
     In recent years, deep learning methods have achieved promising object detection results for ground imagery and outperformed traditional methods.
 Among them,~\glspl{gls:DCNN} have been widely used~\cite{alexnet,Simonyan2015VeryRecognition,resnetHe15}.
In the~\gls{gls:RS} domain, newly introduced large-scale multi-class image datasets such as DOTA~\cite{dota} have provided the opportunity to leverage the applications of deep learning methods.	
The majority of current deep learning-based methods detect objects based on~\glspl{gls:HBB}, which are appropriate for ground-level images.
However, in the~\gls{gls:RS} scenarios, objects can be arbitrarily oriented.
Therefore, utilizing~\glspl{gls:OBB} is highly recommended, especially when multiple objects are located tightly close to each other (\eg cars in parking lots).

    \glspl{gls:RCNN} such as (Fast(er))-\gls{gls:RCNN}~\cite{rcnn, fasterrcnnNIPS2015,fastrcnn} and Mask-\gls{gls:RCNN}~\cite{he2017maskrcnn} have achieved state-of-the-art object detection results in large-scale ground imagery datasets~\cite{Everingham10thepascal,Lin2014}.
 Fast-\gls{gls:RCNN}~\cite{fastrcnn} improves the detection accuracy of~\gls{gls:RCNN}~\cite{rcnn} by using a multi-task loss function for the simultaneous region proposal regression and classification tasks.
 As an improvement, Faster-RCNN integrates an end-to-end trainable network, called~\gls{gls:RPN}, to learn the region proposals for increasing the localization accuracy of Fast-\gls{gls:RCNN}.
 To further improve Faster-RCNN, one could perform multi-scale training and testing to learn feature maps in multiple levels; however, this will increase the memory usage and inference time.
 
 Another alternative is image or feature pyramids~\cite{DBLP:journals/corr/PinheiroLCD16,recombinator,laplacian,stacked, fpn,D-FPN}.
    Recently, Lin et al.~\cite{fpn} proposed the~\gls{gls:FPN} which extracts feature maps through a feature pyramid, thus facilitating object detection in different scales, at a marginal extra cost.
	Although joint image and feature pyramids may further improve results, this is avoided due to its computation cost.

    Object detection in~\gls{gls:RS} images has been investigated by a number of works in the recent years.
    The majority of the proposed algorithms focus on object detection with a small number of classes and a limited range of~\glspl{gls:GSD}.
    Liu and Mattyus~\cite{Lui} proposed~\gls{gls:HOG} features and the AdaBoost method for feature classification to detect multi-class oriented vehicles.
    Although this approach achieves a fast inference time, it does not have high detection accuracy as it lacks high-level feature extraction.
    Sommer et al.~\cite{ies_2017_sommer_deep_learning} and Tang et al.~\cite{Tang2017VehicleDI} proposed \gls{gls:RCNN}-based methods using hard-negative mining together with concatenated and deconvolutional feature maps.
    They showed that these methods achieve high accuracies in single-class vehicle detection in aerial images for~\glspl{gls:HBB} task.
    Liu et al.~\cite{DBLP:journals/corr/abs-1711-09405} proposed rotated region proposals to predict object orientation using~\gls{gls:SSD}~\cite{DBLP:conf/eccv/LiuAESRFB16} improving the localization of the~\glspl{gls:OBB} task.
Yang et al.~\cite{XueYang2018} improved~\cite{DBLP:journals/corr/abs-1711-09405} by integrating~\glspl{gls:FPN}.

	In this paper, we focus on improving the object localization of region-based methods applied to aerial and satellite images.
    We propose a new end-to-end CNN to address the aforementioned challenges of multi-class object detection in~\gls{gls:RS} images. The proposed method is able to handle images with a wide range of scales, aspect ratios,~\glspl{gls:GSD}, and complex backgrounds. In addition, our proposed method achieves accurate object localization by using~\glspl{gls:OBB}. 
   More specifically, the method is composed of the following consecutive modules:~\gls{gls:ICN},~\gls{gls:DIN},~\gls{gls:FPN},~\gls{gls:R-RPN},~\gls{gls:R-ROI}, and~\gls{gls:R-NMS}.
The main contributions of our work are the following:
\begin{itemize}
  \item We propose a new joint image cascade and feature pyramid network (ICN and~\gls{gls:FPN}) which allows extracting information on a wide range of scales and significantly improves the detection results. 
  \item We design a~\gls{gls:DIN} module as a domain adaptation module for adapting the pre-trained networks to the~\gls{gls:RS} domain using deformable convolutions and multi-size convolution kernels.
  \item We propose a new loss function to enforce the detection coordinates, forming quadrilaterals, to shape rectangles by constraining the angles between the edges to be 90 degrees. This augments object localization.
  \item We achieve significant improvements on three challenging datasets in comparison with the state of the art.
\end{itemize}
In addition, we employ rotational region proposals to capture object locations more accurately in~\gls{gls:RS} images.
Finally, in order to select the best localized regions and to remove redundant detections, we apply R-NMS which is the rotational variant of the conventional NMS.
Furthermore, we initialize anchor sizes in~\glspl{gls:R-RPN} with clustered data from rotated ground truth bounding boxes proposed by Redmon and Farhadi~\cite{redmon2017yolo9000} rather than manual initialization used in Faster-RCNN.
In order to evaluate the proposed method, we applied it to the DOTA~\cite{dota} dataset, a recent large-scale satellite and aerial image dataset, as well as the UCAS-AOD and NWPU VHR-10 datasets.
Results show that the proposed method achieves a significantly higher accuracy in comparison with state-of-the-art object detection methods.

	\section{Proposed Method}\label{sec:method}
	Figure~\ref{fig:overflow} gives a high-level overview of our joint horizontal and oriented bounding box prediction pipeline for multi-class object detection. Given an input image, combined image cascade and feature pyramid networks (ICN and~\gls{gls:FPN}) extract rich semantic feature maps tuned for objects of substantially varying sizes.
    Following the feature extraction, a~\gls{gls:R-RPN} returns category-agnostic rotated regions, which are then classified and regressed to bounding-box locations with a~\gls{gls:R-ROI}.
    During training, we minimize a multi-task loss both for~\gls{gls:R-RPN} and~\gls{gls:R-ROI}. 
    To obtain rectangular predictions, we further refine the output quadrilaterals by computing their minimum bounding rectangles. 
Finally,~\gls{gls:R-NMS} is applied as a post-processing.
\overviewfigure
	\subsection{Image Cascade, Feature Pyramid, and Deformable Inception Subnetworks}\label{2-1}
	In order to extract strong semantic information from different scales, this work aims at leveraging the pyramidal feature hierarchy of~\glspl{gls:CNN}.
    Until recently, feature extraction was typically performed on a single scale~\cite{fasterrcnnNIPS2015}. Lately, however, multi-scale approaches became feasible through~\gls{gls:FPN}~\cite{fpn}.
	As argued in~\cite{fpn}, the use of pyramids both at the image and the feature level is computationally prohibitive. Nevertheless, here we show that by an appropriate weight sharing, the combination of~\gls{gls:ICN} (Figure~\ref{fig:icn}) and~\gls{gls:FPN} (Figure~\ref{fig:FPN}) becomes feasible and outputs proportionally-sized features at different levels/scales in a fully-convolutional manner.
	This pipeline is independent of the backbone~\gls{gls:CNN} (\eg AlexNet~\cite{alexnet}, VGG~\cite{Simonyan2015VeryRecognition}, or ResNet~\cite{resnetHe15}). Here, we use ResNet~\cite{resnetHe15}. 
In the \textbf{\gls{gls:ICN}}, as illustrated in Figure~\ref{fig:icn}, we use ResNet to compute a feature hierarchy ${C_1, C_2, C_3, C_4, C_5}$,  which correspond to the outputs of the residual blocks: conv1, conv2, conv3, conv4, and conv5 (blue boxes in Figure~\ref{fig:icn}). The pixel strides for different residual boxes are 2, 4, 8, 16, and 32 pixels with respect to the input image.

	To build our image cascade network, we resize the input image by bilinear interpolation to obtain four scaled versions ($1.5\times, 1\times, 0.75\times, 0.5\times$) and extract the feature hierarchy using ResNet subnetworks. For example, while all five residual blocks are used for the upsampled input ($1.5\times$), for the half-resolution version ($0.5\times$), only $C_4$ and $C_5$ are used.
	The cascade network is thus composed of different subnetworks of the ResNet sharing their weights with each other.
	Therefore, apart from resizing the input image, this step does not add further computation costs with respect to the single resolution baseline. \gls{gls:ICN} allows combining the low-level semantic features form higher resolutions (used for detecting small objects) with the high-level semantic features from low resolutions (used for detecting large objects).
    This helps the network to handle~\gls{gls:RS} images with a wide range of~\gls{gls:GSD}.
    A similar definition of~\gls{gls:ICN} was proposed for real-time semantic segmentation in~\cite{zhao2017icnet}, but without taking into account different scales in the feature domain and using a cascaded label for each level to compensate for the sub-sampling.
    Such a cascaded label is more suitable for semantic segmentation.
	
	\textbf{~\glspl{gls:FPN}}~\cite{fpn} allow extracting features at different scales by combining the semantically strong features (from the top of the pyramid) with the semantically weaker ones (from the bottom) via a top-down pathway and lateral connections (cf.\ Figure~\ref{fig:FPN}).
    The original bottom-up pathway of~\gls{gls:FPN} (\ie the feed-forward computation of the backbone~\gls{gls:CNN}) is here replaced with the feature hierarchy extraction of~\gls{gls:ICN}, more specifically with the output of their residual blocks $C_{i}$, $i\in\{1,2,3,4,5\}$.
    The top-down pathway upsamples coarse-resolution feature maps ($M_i$) by a factor of 2 and merges them with the corresponding bottom-up maps $C_{i-1}$ (\ie the lateral connections).
    The final set of feature maps $P_i$, $i\in\{1,2,3,4,5\}$, is obtained by appending $3\times3$ convolutions to $M_i$ to reduce the aliasing effect of upsampling.
    We refer the reader to the work of Lin et al.~\cite{fpn} for more details on FPNs.
\cascadefigure
\fpmidnfigure
In the original FPN, the output of each $C_i$ goes through a $1\times1$ convolution to reduce the number of feature maps in $M_i$.    
Here, we replace the $1\times1$ convolution with a \textbf{\gls{gls:DIN}} (Deformable Inception Network, cf.\ Figure~\ref{fig:FPN}) to enhance the localization properties of~\glspl{gls:CNN}, especially for small objects which are ubiquitous in~\gls{gls:RS} datasets.
Although Inception modules~\cite{inception} have shown promising results in various tasks such as object recognition, their effectiveness for  detection has not been extensively studied. 
While most current state-of-the-art methods, such as Faster-RCNN, R-FCN~\cite{NIPS2016_6465}, YOLOv3~\cite{redmon2017yolo9000}, and~\gls{gls:SSD}~\cite{DBLP:conf/eccv/LiuAESRFB16}, focus on increasing the network depth, the benefit of Inception blocks lies in capturing details at varied scales which is highly desirable for~\gls{gls:RS} imagery.

Deformable networks aim at overcoming the limitations of~\glspl{gls:CNN} in modeling geometric transformations due to their fixed-size convolution kernels.
When applying the models pretrained on ground imagery (such as our ResNet backbone) to~\gls{gls:RS} images, the parameters of traditional convolution layers cannot adapt effectively to the new views of objects leading to degradations in localization performance.
Using deformable convolutions in~\gls{gls:DIN} helps accommodating such geometric transformations~\cite{dai17dcn}.
Furthermore, the offset regression property of deformable convolution layers helps localizing the objects even outside the kernel range. Here, we train the added offset layer from scratch to let the network adjust to the new domain.
$1\times1$ convolution layers reduce dimensions by half for the next deformable convolution (def-conv) layers.
The channel input to~\gls{gls:DIN} is divided equally among the four~\gls{gls:DIN} branches.
In our experiments, we did not observe an improvement by using $5\times5$ def-conv layers, hence the use of $3\times3$ layers.
	
	\subsection{Rotation Region Proposal Network (R-RPN)}
	The output of each $P_i$ block in the~\gls{gls:FPN} module is processed by multi-scale rotated region proposal networks (R-RPN) in order to provide rotated proposals, inspired by~\cite{Jianqi17RRPN}.
	More precisely, we modify~\gls{gls:RPN} to propose rotated regions with 0, 45, 90, and 135 degrees rotation, not differentiating between the front and back of objects.
    For initializing the anchors, we cluster the scales and aspect ratios using K-means++ with the~\gls{gls:iou} distance metric~\cite{redmon2017yolo9000}.
	We  assign anchors with four different orientations to each level, $P_2$ through $P_6$\footnote{$P_6$ is a stride 2 sub-sampling of $P_5$ used to propose regions for large objects. $P_1$ is not computed due to its large memory footprint.}.
	As in the original~\gls{gls:RPN}, the output feature maps of~\gls{gls:FPN} go through a $3\times3$ convolutional layer, followed by two parallel $1\times1$ fully-connected layers: an objectness classification layer ($obj$) and a box-regression layer ($reg$) (cf.\ Figure~\ref{fig:overflow}).
	For training, we assign labels to the anchors based on their~\glspl{gls:iou} with the ground-truth bounding boxes.
	In contrast to the traditional~\gls{gls:RPN}, we use the smooth $l_1$ loss to regress the four corners $(x_i, y_i)$, $i\in\{1, 2, 3, 4\}$, of the~\gls{gls:OBB} instead of the center point $(x, y)$, and size ($w$ and $h$) of the~\gls{gls:HBB}. 
    In this case, $(x_1, y_1)$ indicates the front of objects which allows to infer their orientations.
    As in Faster-RCNN, we minimize the multi-task loss
	\begin{align}
    L\left( \{p_i\},\{t_i\} \right) &= 
      \frac{1}{N_{obj}}\sum_{i} L_{obj}({p_i},{p_i^*}) 
      + \lambda \frac{1}{N_{reg}} \sum_{i}{p_i^*} L_{reg}\left( {t_i},{t_i^*} \right) 
      \quad,
    \label{eq:RRPN-2} 
	\end{align}
where, for an anchor $i$ in a mini-batch, $p_i$ is its predicted probability of being an object and ${p_i^*}$ is its ground-truth binary label. 
For classification (object/not-object), the log-loss $L_{obj}({p_i},{p_i^*}) =  - {{p_i^*}} \log{p_i}$ is used, while we employ the smooth $l_1$ loss
\begin{align}
  L_{reg}({t_i},{t_i^*}) &= l_1^{\text{smooth}}({t_i} - {t_i^*}) 
  &\text{with }   
  l_1^{\text{smooth}}(x) &=  
  \begin{cases}
    0.5x^2        & \text{if }  |x| < 1\\
    |x|-0.5       & \text{otherwise}
  \end{cases}
  \label{eq:smoothl1}
\end{align}
for the bounding box regression. Here, 
\begin{align}
	t_{xi}   &= (x_i - x_{i,a})/w_a, &   t_{yi}   &= (y_i - y_{i,a})/h_a \\
	t_{xi}^* &= (x_i^* - x_{i,a})/w_a, & t_{yi}^* &= (y_i^* - y_{i,a})/h_a 
\end{align}
are the four parameterized coordinates of the predicted and ground-truth anchors with $x_i$, $x_{i,a}$, and $x_{i}^*$ denoting the predicted, anchor, and ground-truth, respectively (the same goes for $y$); and $w_a$ and $h_a$ are width and height of the anchor.
    $N_{obj}$ and $N_{reg}$ are normalizing hyper-parameters (the mini-batch size and number of anchor locations); and $\lambda$ is the balancing hyper-parameter between the two losses which is set to 10.
	
	\subsection{Rotated Region of Interest Network (R-ROI)}
	Similar to~\cite{fpn}, we use a multi-scale ROI pooling layer to process the regions proposed by R-RPN.
    Because the generated proposals are rotated, we rotate them to be axis-aligned.
    The resulting fixed-length feature vectors are fed into sequential fully-connected (\emph{fc}) layers, and are finally sent through four sibling \emph{fc} layers, which -- for each object proposal -- output the class prediction, refined HBB and~\gls{gls:OBB} positions, as well as the angles of~\glspl{gls:OBB}. 
	
	As seen for~\glspl{gls:R-RPN},~\glspl{gls:OBB} are not restricted to be rectangular:~\gls{gls:R-RPN} predicts the four corners of quadrilaterals without any constraint on the corners or edges. 
	However, we observed that annotators tend to label rotated objects in~\gls{gls:RS} images with quadrilaterals that are close to rotated rectangles.
	In order to enforce a rectangular shape of~\glspl{gls:OBB}, we propose a new loss  that considers the angles between adjacent edges, \ie we penalize angles that are not $90^{\circ}$.
    
	Let us consider $P_{ij}$ a quadrilateral side connecting the corners $i$ to $j$, where $i,j\in\{1, 2, 3, 4\}$ and $i\neq j$.
	Then, using the cosine rule, we calculate the angle between adjacent sides (\eg $\theta_1$ between $P_{12}$ and $P_{13}$) as: 
	\begin{equation}
	\theta_1 = \arccos((|P_{12}|^2 + |P_{13}|^2 - |P_{23}|^2) / (2 * |P_{12}| * |P_{13}|))\quad,
	\end{equation}
	where $|P_{ij}|$ is the length of the side $P_{ij}$.
	There are multiple ways to constrain $\theta_l, l\in\{1, 2, 3\}$ to be right angles. (Note that $\theta_4$ can be computed from the other three angles). We experimented with the following three angle-losses:
    \begin{equation}
    \begin{split}
	\text{Tangent L1}: L_{angle-OBB}(\theta)&= \sum_{l=1}^{3}(|tan(\theta_l - 90)|)\\
    \text{Smooth L1}: L_{angle-OBB}(\theta) &= \sum_{l=1}^{3}smooth_{L1}(|\theta_l - 90|) \\
	\text{L2}: L_{angle-OBB}(\theta) &= \sum_{l=1}^{3}\left\lVert (\theta_l - 90) \right\rVert^2.
    \end{split}
	\end{equation}
	Our final loss function is a multi-task loss  composed of four losses that simultaneously predict the object category ($L_{cls}$), regress both~\gls{gls:HBB} and~\gls{gls:OBB} coordinates ($L_{loc-HBB}$ and $L_{loc-OBB}$), and enforce~\glspl{gls:OBB} to be rectangular ($L_{angle-OBB}$):
	\begin{equation}
	\begin{split}
	L(p,u,t^u,v) = L_{cls}(p,u) +  \lambda[u\geq1]L_{loc-HBB}(t^u,v) + \\
	\lambda[u\geq1]L_{loc-OBB}(t^u,v) + \lambda[u\geq1]L_{angle-OBB}(\theta)\quad,
	\end{split}
	\end{equation}
	where $L_{cls}({p},{u}) =  - {u} \log{p}$ and $L_{loc-OBB}(t^u,v)$ is defined similar to $L_{reg}$ as in ~\gls{gls:R-RPN} above.
    $u$ is the true class and $p$ is the discrete probability distribution for the predicted classes, defined over $K+1$ categories as $p= (p_0, ...., p_K)$ in which ``1'' is for the background category.
	$t^u = (t^u_{xi},t^u_{yi})$ is the predicted~\gls{gls:OBB} regression offset for class $u$ and $v = (v_{xi}, v_{yi})$ is the true~\gls{gls:OBB} ($i\in\{1, 2, 3, 4\}$).
    $L_{loc-HBB}(t^u,v)$ is defined similar to $L_{reg}$ in Faster-RCNN in which instead of~\gls{gls:OBB} coordinates, $\{xmin, ymin, w, h\}$ (the upper-left coordinates, width and height) of $t^u$ and $v$ for the corresponding~\gls{gls:HBB} coordinates are utilized.
	In case the object is classified as background, $[u\geq1]$ ignores the offset regression.
    The balancing hyper-parameter $\lambda$ is set to $1$.
	To obtain the final detections, we compute the minimum bounding rectangles of the predicted quadrilaterals.
    As the final post-processing, we apply~\gls{gls:R-NMS} in which the overlap between rotated detections is computed to select the best localized regions and to remove redundant regions.
    	\section{Experiments and Discussion}\label{sec:experiments}
	In this section, we present and discuss the evaluation results of the proposed method on three~\gls{gls:RS} image datasets.
    All experiments were conducted using NVIDIA Titan X GPUs.
The backbone network's weights were initialized using the ResNet-50/101 and ResNeXt-101 models pretrained on ImageNet~\cite{Deng2009}.
	Images were preprocessed as described in baseline~\cite{dota}. Furthermore, the learning rate was 0.0005 for 60 epochs with the batch size of 1 using flipped images as the data augmentation.
    Additionally, during training, we applied~\gls{gls:OHEM}~\cite{AbhinavShrivastava2016} to reduce false positives and we use Soft-NMS~\cite{softnms} as a more accurate non-maximum suppression approach only for the HBB benchmark.
		\subsection{Datasets}
	The experiments were conducted on the DOTA~\cite{dota}, UCAS-AOD~\cite{Zhu2015OrientationRO}, and NWPU VHR-10~\cite{Chengnwhu} datasets which all have multi-class object annotations. 
    
\textbf{DOTA} is the largest and most diverse published dataset for multi-class object detection in aerial and satellite images.
	It contains 2,806 images from different camera sensors,~\glspl{gls:GSD} (\SI{10}{\cm} to \SI{1}{\m}), and sizes to reflect real-world scenarios and decrease the dataset bias. The images are mainly acquired from Google Earth, and the rest from the JL-1 and GF-2 satellites of the China Center for Resources Satellite Data and Application.
    Image sizes vary from 288 to 8,115 pixels in width, and from 211 to 13,383 pixels in height.
	There are 15 object categories: plane, baseball diamond (BD), bridge, ground field track (GTF), small vehicle (SV), large vehicle (LV), tennis  court (TC), basketball court (BC), storage tank (SC), soccer ball field (SBF), roundabout (RA), swimming  pool (SP), helicopter (HC), and harbor.
	DOTA is split into training (1/2), validation (1/6), and test (1/3) sets.
	
\textbf{UCAS-AOD} contains 1,510 satellite images ($\approx$ \SI{700 x 1300}{\px}) with 14,595 objects annotated by~\glspl{gls:OBB} for two categories: vehicles and planes. The dataset was randomly split into 1,110 training and 400 testing images.
    
\textbf{NWPU VHR-10} contains 800 satellite images ($\approx$ \SI{500 x 1000}{\px}) with 3,651 objects were annotated with~\glspl{gls:HBB}.
There are 10 object categories: plane, ship, storage tank, baseball diamond, tennis court, basketball court, ground track field, harbor, bridge, and small vehicle.
For training, we used non-rotated~\gls{gls:RPN} and~\gls{gls:ROI} networks only for the~\glspl{gls:HBB} detection task.
    
    \subsection{Evaluation}\label{sec:evaluations}
	In order to assess the accuracy of our detection and the quality of region proposals, we adapted the same~\gls{gls:map} and~\gls{gls:ar} calculations as for DOTA~\cite{dota}.
	We conducted ablation experiments on the validation set of DOTA.
	Furthermore, we compare our method to the ones in~\cite{dota} for~\gls{gls:HBB} and~\gls{gls:OBB} prediction tasks as well as Yang et al.~\cite{XueYang2018} for~\gls{gls:OBB} task based on the test set whose ground-truth labels are undisclosed.
	The results reported here were obtained by submitting our predictions to the official DOTA evaluation server\footnote{\url{http://captain.whu.edu.cn/DOTAweb/evaluation.html}}.
  We used $0.1$ threshold for~\gls{gls:R-NMS} and $0.3$ for Soft-NMS.
  
	\textbf{The impact of~\gls{gls:ICN}:}
	From Table~\ref{tab:cascade} shows the evaluation results of~\gls{gls:ICN}. According to the table, adding~\gls{gls:OHEM} to ResNet-50 improved the accuracy by a narrow margin.
    Using a deeper network such as ResNet-101 further improved the accuracy.
    As a next step, adding a $1.5\times$ cascade level increased~\gls{gls:map} by around 2\% indicating that the up-sampled input can have a significant impact.
    Based on this, we added smaller cascade levels such as $0.75\times$ and $0.5\times$, which however, increased the accuracy to a lesser extent.
    This could be due to the fact that the majority of  objects within this dataset are small, so reducing resolution is not always optimal.
    Further increasing the cascade levels (\eg $1.75\times$ and $2\times$) degraded the accuracy, which is due to the lack of annotations for very small objects such as small vehicles.
    We argue that extracting ResNet features on upsampled images ($1.5\times$) is beneficial for the small objects in the DOTA dataset, whereas doing this on the downsampled input ($0.75\times, 0.5\times$) brings smaller improvements because of the lower number of large objects in the dataset.
	We observed that replacing ResNet-101 with ResNeXt-101 causes a small drop in accuracy which could be due to the shallower architecture of ResNeXt-101.
	Results indicated that using a higher number of proposals (2000) increases the accuracy to a small degree, which however came with an increased computation cost; thus, we considered 300 proposals for the rest of our experiments.
    
  \icntable
	\textbf{The impact of~\gls{gls:DIN}:}
	From Table~\ref{tab:din} we see that replacing the $1\times1$ convolution after the residual blocks $C_i$ by~\gls{gls:DIN} can augment mAP by more than 2\%.
	More specifically, using~\gls{gls:DIN} after lower level $C_i$s resulted in slightly higher accuracy than using it after higher levels (\eg mAP for C4 $>$ mAP for C5). In addition, employing~\gls{gls:DIN} after multiple $C_i$s can further improve model performance (\eg mAP for C4 $<$ mAP for C4---C5 $<$ mAP for C3---C5). 
    Kernel size strongly affects the high resolution (semantically weak) features. Thus, applying~\gls{gls:DIN} to the low-level $C_i$s enriched the features and adapts them to the new data domain.
    Comparing the last two rows of Table~\ref{tab:din}, we see that deformable convolutions also have a positive impact; however, the improvement is smaller.
    
\dintable
\dotaclassesfigure
\dotavehiclesfigure
	\textbf{Rotated~\gls{gls:RPN} and~\gls{gls:ROI} modules:}
    Using clustered initialized anchors with rotation, we obtained an additional $0.7\%$ mAP.
    To initialize anchors, we selected 18 anchors compared to 15 in Faster-RCNN in clustering ground-truth~\glspl{gls:OBB}.
	We observed no significant increase in~\gls{gls:iou} with higher number for anchors.
	Furthermore, we considered each anchor at four different angles (0, 45, 90, 135 degrees rotation).
    The total number of anchors is thus $18\times4$.
    Table~\ref{tab:r} shows that using rotated proposals in the~\gls{gls:R-RPN}/~\gls{gls:R-ROI} layers improves~\gls{gls:map} by 1.4\%, indicating that these proposals are more appropriate for~\gls{gls:RS} images.
    
	In addition, we see that using a joint loss function (for~\gls{gls:HBB} and~\gls{gls:OBB} prediction) can increase the prediction of OBBs by 0.81\%~\gls{gls:map}.
    We believe that~\glspl{gls:HBB} provide useful ``hints'' on the position of the object for regressing~\glspl{gls:OBB} more accurately.
    This is not the case for~\gls{gls:HBB} prediction: here, using only the~\gls{gls:HBB} regression loss achieves 3.98\% higher~\gls{gls:map} as compared to the joint loss. This could be due to the complexity that~\gls{gls:OBB} imposes on the optimization problem.
    Thus, we apply our algorithm on the HBB benchmark without the~\gls{gls:OBB} loss.
	
	\textbf{Enforcing rectangular bounding boxes:} We investigated three different loss functions to enforce the rectangularity of the quadrilateral bounding boxes.
	Results in Table~\ref{tab:r} show that all three angle losses improve the output accuracy and angle L2 performs the best.
		The reason behind the lower performance of angle tangent L1 could be the property of the $tangent$ function: it leads to very high loss values when the deviation from the right angle is large.
	Angle smooth L1 performs marginally worse than angle L2 which could be due to its equal penalization for deviations larger than 1 degree from the right angle.
\rpnroitable

    By studying the recall-IoU curve, we noticed that very small and very large objects (\eg small vehicles and very large bridges) have the lowest localization recall and medium-size objects have the highest recall.
    Overall~\gls{gls:ar} for the proposals on DOTA is 61.25\%.
    A similar trend is observed for prec-recall curves.
    
\ucasnwpfigure
\fpanalysisfigure
\HBBbenchmarktable
   \textbf{On False Positives:} To investigate false positives, we used the object detection analysis tool from~\cite{DerekHoiem2012}.
    For the sake of brevity, we merge the bridge and harbor as the long objects class, and the LV, SV, and ship classes as the vehicles class.
    Similar observations were made for the rest of the classes. 
	The large blue area in Figure~\ref{fig:fps} indicates that our method detects object categories with a high accuracy.
	Moreover, recall is around 80\% (the red line) and is even higher with ``weak'' (10\% overlap with the ground truth) localization criteria (dashed red line).
	The majority of confusions are with the background (the green area) while the confusion with similar object classes is much smaller (yellow area).
	This issue is more severe for long objects.
    Although using only down-sampled levels in the image cascade alleviates this issue, it lowers the performance for small objects.
    Since the proposals are not able to capture long objects effectively, they cause a large localization error.
	Additionally, the false positives for similar-classes often occur for vehicles: small and large vehicles are mistaken for each other.
    
	\textbf{Comparison with the state of the art:}
	Tables~\ref{tab:benchmarkhbb}~and~\ref{tab:benchmarkobb} show the performance of our algorithm on the~\gls{gls:HBB} and~\gls{gls:OBB} prediction tasks DOTA,
    based on the official evaluation of the methods on the test set with non-disclosed ground-truth.
    We evaluate our method in two scenarios: training only on the `train' subset, and training on the training and validation sets (`trainval').
    Our method significantly outperforms all the published methods evaluated on this benchmark, and training on `trainval' brings an additional 2-4\% in~\gls{gls:map} over training only on `train'.
    Looking at individual class predictions, only the~\glspl{gls:map} of the helicopter, bridge, and SBF classes are lower than the baseline, possibly due to their large (and unique) size, complex features, and low occurrence in the dataset.
    
\OBBbenchmarktable
    \textbf{Generalization on the NWPU VHR-10 and UCAS-AOD datasets:} As shown in Table~\ref{tab:generalization}, our algorithm significantly improves upon the baseline also on these two additional datasets.
This demonstrates the good generalization capability of our approach.
Results are competitive even when we trained our algorithm only on DOTA dataset.
\ucasnwptable
	\section{Conclusions}\label{sec:conclusion}
	In this work, we presented a new algorithm for multi-class object detection in unconstrained~\gls{gls:RS} imagery evaluated on three challenging datasets.
	Our algorithm uses a combination of image cascade and feature pyramids together with rotation proposals.
	We enhance our model by applying a novel loss function for geometric shape enforcement using quadrilateral coordinates.
    Our method outperforms other published algorithms~\cite{dota,XueYang2018} on the DOTA dataset by a large margin.
    Our approach is also robust to differences in spatial resolution of the image data acquired by various platforms (airborne and space-borne).
\bibliographystyle{splncs04}
\bibliography{egbib}
\end{document}